# Looks can be Deceptive: Distinguishing Repetition Disfluency from Reduplication

Arif Ahmad[1], Mothika Gayathri Khyathi[1], Pushpak Bhattacharyya[1]

[1]CFILT, Indian Institute of Technology Bombay
`190110010@iitb.ac.in`, `khyathimothika3@gmail.com`, `pb@cse.iitb.ac.in`

## Abstract

Reduplication and repetition, though similar in form, serve distinct linguistic purposes. Reduplication is a deliberate morphological process used to express grammatical, semantic, or pragmatic nuances, while repetition is often unintentional and indicative of disfluency. This paper presents the first large-scale study of reduplication and repetition in speech using computational linguistics. We introduce IndicRedRep, a new publicly available dataset containing Hindi, Telugu, and Marathi text annotated with reduplication and repetition at the word level. We evaluate transformer-based models for multi-class reduplication and repetition token classification, utilizing the Reparandum-Interregnum-Repair structure to distinguish between the two phenomena. Our models achieve macro F1 scores of up to 85.62% in Hindi, 83.95% in Telugu, and 84.82% in Marathi for reduplication-repetition classification.

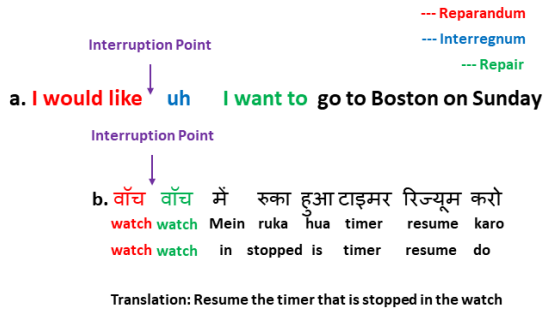

Figure 1: Examples showing the four regions of any disfluency: Redarandum, Interruption Point, Interregnum, and Repair. Not all parts are necessary to be present in every example of a disfluency; as can be seen in Example (b) in the Figure, with no interregnum.

## 1 Introduction

Speech recognition technology significantly enhances accessibility for individuals with disabilities by improving their interaction with technology and overall quality of life (Green et al., 2021; MacArthur and Cavalier, 2004; Noyes and Frankish, 1992). It also plays a vital role in consumer electronics, where Automatic Speech Recognition (ASR) technologies like Siri, Google Assistant, and Amazon Alexa enrich user experiences through voice-activated controls (Buteau and Lee, 2021; Juang and Rabiner, 2005).

Research shows that speech disfluencies, such as repetitions, can notably increase Word Error Rates (WER) by up to 15% (Goldwater et al., 2008). Addressing these disfluencies in ASR systems can improve performance, as demonstrated by enhancements in Machine Translation (MT) systems' BLEU scores (Cho et al., 2014). This paper focuses on repetition—a type of disfluency characterized by the unintended recurrence of words or phrases, which typically occurs during moments of cognitive processing, such as recalling a word or structuring a thought (Tree, 1995).

Interestingly, repetition shares structural similarities with reduplication—a deliberate linguistic process used globally to alter word meanings, indicating attributes like plurality or intensity. While both processes involve repetition, their functions and implications differ significantly, with reduplication playing a grammatical and semantic role in languages and repetition often marking interruptions in speech flow (Newman, 2000; Bauer, 2003; Xu, 2012; Kajitani, 2005).

Given the prevalence of both reduplication and repetition in spontaneous speech, and their significant implications for language technologies, this study introduces a novel dataset named "**IndicRedRep**". This dataset fo-

| Language | Word (Meaning) | Reduplicated Word (Meaning) |
|---|---|---|
| Indonesian/Malay | orang (person) | orang-orang (people) |
| Tagalog | bili (buy) | bili-bili (to buy here and there) |
| Tamil | kaal (leg) | kaal-kaal (legs) |
| Punjabi | xushii (happy) | xushii-xushii (happily) |
| Mandarin Chinese | 妈 (mā, mother) | 妈妈 (māma, mommy) |
| Hawaiian | wiki (quick) | wiki-wiki (very quick) |
| Samoan | pili (cling) | pili-pili (to cling repeatedly) |
| Turkish | ev (house) | ev-ev (every house) |

Table 1: Examples of Morphological Reduplication in Various Languages Demonstrating Pluralization, Intensification, and Other Grammatical or Semantic Changes

cuses on these phenomena in multiple Indic languages. Research indicates that disfluencies, including reduplication and repetition, can constitute up to 5.9% of words in spontaneous speech, with repetitions accounting for over half of these disfluencies (Godfrey et al., 1992; Shriberg, 1996). IndicRedRep aims to facilitate the development of models capable of distinguishing between reduplication and repetition, treating it as a sequence labeling problem

The contributions of this work are summarized below:

- Creation of "IndicRedRep," a new dataset available to the public that includes over 4.5K Hindi, 1.6K Telugu, and 1.6K Marathi sentences, all annotated with labels for reduplication and repetition. This is the first dataset of its kind to offer token-level annotations for these features in any language.

- Development and implementation of a novel methodology using the Reparandum-Interregnum-Repair (RiR) structure, which has improved the macro F1 score by 2% across all tested languages.

- Empirical evaluation of various classical sequence labeling models and transformer-based models on the new dataset to classify reduplication and repetition accurately.

- Detailed linguistic analysis of the dataset across three languages—Hindi, Telugu, and Marathi—to understand the unique challenges and behaviors of models when dealing with different linguistic contexts.

## 2 Related Work

Reduplication and repetition are well-studied phenomena in the domains of morphology and speech disfluencies, respectively. The related work is organized into two main themes: a) computational models and frameworks developed for processing reduplication in multiword expressions, b) approaches and models employed for detecting and addressing repetition as speech disfluency. Here, we discuss the literature most relevant to our work and establish the relationship between them.

### 2.1 Reduplication as Multiword Expression

Multiword expressions (MWEs) are a cornerstone of linguistic studies and pose significant challenges in natural language processing (NLP) due to their complex, non-compositional nature. Recent research highlights a framework for integrating MWE processing into NLP systems to improve linguistic understanding (Baldwin and Kim, 2010; Sag et al., 2002).

Significant efforts have been made to computationally address reduplication across languages such as Bengali, Cantonese, Mandarin Chinese, Indonesian, Sanskrit, Hindi, and Marathi (Chakraborty and Bandyopadhyay, 2010; Lam, 2013; Chen et al., 1992; Mistica et al., 2009; Kulkarni et al., 2012; Singh et al., 2016). Notable computational models include the use of two-way finite-state transducers (2-way FSTs) and finite-state buffered machines (FSBMs), which effectively model reduplicative processes (Dolatian and Heinz, 2018; Wang, 2021). Additionally, integrating reduplicated expressions into machine translation has enhanced translation accuracy (Doren Singh and Bandyopadhyay, 2011). The creation of the RedTyp database marks a significant advancement in the cataloging of reduplicative morphemes, aiding both theoretical and computational studies (Dolatian and Heinz, 2019). While these studies offer significant theoretical insights, no previous work has released a large-scale dataset specifically for the study of reduplication and repetition.

## 2.2 Repetition as Speech Disfluency

Repetition is a well-known speech disfluency often observed in spontaneous and unscripted speech (Shriberg, 1994). It refers to the unintentional recurrence of words, phrases, or sounds, which may occur due to hesitations, corrections, or cognitive processing.

It is tackled using various computational techniques aimed at enhancing speech recognition and processing. These techniques include Sequence Tagging, Parsing-based, and Noisy Channel models, each leveraging different aspects of machine learning and syntactic analysis (Liu et al., 2006; Georgila et al., 2010; Ostendorf and Hahn, 2013; Zayats et al., 2016, 2014; Ferguson et al., 2015; Wang et al., 2018, 2020; Qian and Liu, 2013; Hough and Schlangen, 2015; Rasooli and Tetreault, 2013; Honnibal and Johnson, 2014; Wu et al., 2015; Yoshikawa et al., 2016; Jamshid Lou and Johnson, 2020; Johnson and Charniak, 2004; Zwarts and Johnson, 2011; Jamshid Lou and Johnson, 2017). Inspired from these works, we move forward with sequence tagging based modeling as this approach has its merits of allowing direct and explicit tagging of disfluencies at the word level, which enables fine-grained detection and classification, critical for developing robust speech recognition systems.

## 3 Background and Definitions

In this section, we define reduplication and repetition, discussing their roles in language and speech. Understanding these definitions is essential for recognizing the differences between these two linguistic phenomena, which is a key focus of this study.

### 3.1 Reduplication.

Reduplication is a morphological process in which a part or the entirety of a word's phonological material is systematically repeated, carrying semantic or grammatical significance. This mechanism is prevalent across numerous global languages, serving diverse linguistic purposes including (plurality, distribution, intensity, aspect (continued or repeated occurrence), reciprocity and more. (Rubino, 2005; Spaelti, 1997).

Examples of complete reduplication in Hindi sentences:

1. आपका       बहुत बहुत    शुक्रिया
   aapka      bohot bohot  shukriya
   Your       very very    thanks

   Translation: Thank you very much.

2. जल्दी जल्दी    काम    खतम    करो
   jaldi jaldi   kaam   khatam  karo
   quickly quickly work  finish  do

   Translation: Finish your work quickly.

3. క్రికెట్    ఆడి ఆడి      ఆయాసం     అనిపిస్తుంది
   cricket   aadi aadi   aayasam    anipisthundi
   cricket   play play   tiredness  feeling

   Translation: I feel tired after playing cricket.

4. क्रिकेट    खेळत खेळत      थकलो
   cricket   khelt khelt    thaklo
   cricket   play play      tired

   Translation: I'm tired after playing cricket.

In these examples, the complete repetition of the base word adds emphasis and intensity to the action or state described, enhancing the overall meaning of the sentences.

In this study, we focus only on full or total reduplication, as this is the case that is confused with repetition. So, from here on whenever we discuss about reduplication, it will mean full reduplication.

### 3.2 Repetition

**Repetition** is a type of Speech Disfluency. Speech Disfluencies are geneally defined as phenomena that interrupt the flow of speech and do not add propositional content to an utterance. Honal and Schultz (2003) describes five types of disfluencies, including Flase Start, Repetition, Editing Term, Filler Word, Interjection. In this paper, we focus on the repetition type of disfluency. Repetition, refers to the unintentional recurrence of whole words, phrases, or segments during spontaneous speech. This form of disfluency often occurs when speakers are trying to recall a word, grappling with a complex thought, or deciding how to phrase something (Tree, 1995).

Examples of word repetition disfluencies:

1. मैं-मैं       घर      जा रहा हूँ।
   mai-mai    ghar    ja raha hoon
   I-I        home    am going

   Translation from Hindi: I-I am going home.

2. वह मेरा      दोस्त दोस्त     है।
   vah mera   dost dost     hai
   He my      friend friend  is

   Translation from Hindi: He is my friend friend.

3. | క్రికెట్ క్రికెట్ | ఆడి | ఆయాసం | అనిపిస్తుంది |
   |---|---|---|---|
   | cricket cricket | aadi | aayasam | anipisthundi |
   | cricket cricket | play | tiredness | feeling |

   Translation from Telugu: I feel tired after playing cricket.

4. | क्रिकेट | खेलत | थकलो थकलो |
   |---|---|---|
   | cricket | khelt | thaklo thaklo |
   | cricket | play | tired tired |

   Translation from Marathi: I'm tired after playing cricket.

In these examples, the repetition of the word does not hold any semantic meaning. Thus examples here are considered an error and hence classified as repetition, unlike examples from Section 3.1.

## 4 IndicRedRep Dataset

This section discusses the formation of the IndicRedRep dataset, which includes data collection, annotation, and key statistics across three Indic languages: Hindi, Marathi, and Telugu, focusing on token-level labels for reduplication and repetition. Hindi resources are more plentiful, necessitating different collection strategies compared to Marathi and Telugu.

### 4.1 Data Collection

To the best of our knowledge, there currently exists no dataset explicitly annotated for both reduplication and repetition. We employed the GramVaani (GV) corpus, a spontaneous telephone speech corpus in Hindi, to establish the Hindi subset of the dataset, addressing the lack of datasets annotated for reduplication and repetition(Deekshitha et al., 2022). For Marathi and Telugu, where similar datasets are absent, we extrapolated from the Hindi data using the Gemma Instruction Tuned models(Team et al., 2024) for sentence generation and engaged native speakers for manual creation of test sets.

It was important to use a dataset containing spontaneous speech rather than read speech, as disfluencies are more commonly observed in spontaneous speech. However, in datasets such as the **Shrutilipi corpus** (Bhogale et al., 2023), **Indian Language Corpora** (Abraham et al., 2020), and **Mozilla Common Voice** (Ardila et al., 2020), which predominantly feature read speech, the majority of word duplications are the result of either reduplication or transcription errors. True instances of repetition were significantly rarer in these sources.

### 4.2 Annotation and Quality Control Process

The collected data was annotated by three trained linguists, who are language majors with a focus on Hindi. They noted the poor quality and many errors in the transcripts of the GramVaani (GV) corpus. Consequently, data annotation occurred in two stages: initially correcting the speech transcripts and subsequently labeling the tokens as reduplication, repetition, or other.

Given the noisy annotations in the GV corpus, which often skipped repetition words, we filtered the corpus to select sentences likely containing reduplication or repetition. After filtering, these sentences were reannotated to properly account for missed instances. This resulted in a clean, ground truth dataset with real-world instances of reduplication and repetition in Hindi, which were then labeled at the token level. For Marathi and Telugu, we used the cleaned and annotated Hindi subset to bootstrap sentence generation with the Gemma Instruction Tuned models (Team et al., 2024). Native speakers of Marathi and Telugu were later engaged to manually create sentences for the test sets, using the Hindi data as a reference.

Annotation guidelines were developed based on existing works (Murthy et al., 2022) and are detailed in Appendix B. To ensure consistency, interannotator agreement was assessed using Fleiss' kappa, resulting in an agreement level of 83.29%, indicating substantial consistency. Quality control included independent re-annotation by multiple annotators, with discrepancies resolved during regular meetings (Sabou et al., 2014). Additionally, the Gemma prompting details used for generating Marathi and Telugu sentences are included in Appendix C.

### 4.3 Dataset Statistics

The GramVaani corpus, inherently rich in colloquial expressions and spontaneous speech patterns, provided an ideal foundation for our specific annotations. Table 2 shows the num-ber of sentences and words, across each split in the dataset. As showcased in Table 3, our annotated dataset comprises of labels: *reduplication*, *repetition* and *other*. The presence of 3,263 instances of repetition and 2,340 of repetition underscores the diversity and richness of this corpus in capturing these linguistic phenomena. Additionally, there are 586 tokens labeled as 'other', indicating elements that did not fit into the primary categories but were distinct from the neutral 'O' category.

### 4.4 Data Splits

The data was divided into training, validation, and test sets following the standard 80:10:10 ratio. The splits were stratified to ensure that the distribution of reduplication and repetition instances was similar across all subsets as can be seen in Table 3.

| Language | Data Splits | #sentences | #words | Split Size |
|---|---|---|---|---|
| Hindi | Training | 3622 | 103602 | 80% |
| | Validation | 453 | 12950 | 10% |
| | Test | 453 | 12950 | 10% |
| Telugu | Training | 1289 | 36860 | 80% |
| | Validation | 161 | 4608 | 10% |
| | Test | 161 | 4608 | 10% |
| Marathi | Training | 1322 | 37822 | 80% |
| | Validation | 165 | 4728 | 10% |
| | Test | 165 | 4728 | 10% |

Table 2: Dataset statistics across three languages for a token classification task

| | Training | Validation | Test | Total |
|---|---|---|---|---|
| repetition | 2598 | 335 | 330 | 3263 |
| reduplication | 1875 | 230 | 235 | 2340 |
| other | 462 | 62 | 62 | 586 |
| **Total** | 4935 | 627 | 627 | 6189 |

Table 3: Number of labels of each type in Training, Validation, Test splits for the **IndicRedRep** dataset

## 5 Modelling

In this section, we detail our approach to differentiate between reduplication and repetition, with a particular focus on utilizing the Reparandum-Interregnum-Repair (RiR) structure. We believe that by considering the context surrounding the repeated elements, we can disambiguate intricate cases where reduplication and repetition coexist. This is also supported by analisysis of the disfluency structure by Shriberg (1994), which we discuss in detail here.

### 5.1 RiR Structure

Shriberg (1994) describes the structure of disfluencies, to consist of four parts: Redarandum, Interruption Point, Interregnum, and Repair. Figure 1 shows an eaxmple of a disfluency following this structure in English as well as in Hindi. Interruption Point is equivalent to the "moment of interruption" and is not explicity present in the transcript, but a part of the speech signal. Hence, we donot capture it in our modelling strategy.

The Reparandum-Interregnum-Repair structure serves as the foundation of our classification methodology. It captures the distinctive patterns associated with reduplication and repetition:

- **Reparandum**: This is the initial segment of the utterance, containing the word or phrase that undergoes repetition or reduplication.

- **Interregnum**: The interregnum is an optional segment that may exist between the repetitions or reduplications. It plays a crucial role in distinguishing the two phenomena, as it often contains disfluent elements or markers.

- **Repair**: The repair segment is where the repetition or reduplication occurs. This is the part of the utterance that replicates the Reparandum, potentially with variations.

### 5.2 Importance of the RiR Structure

Our motivation for incorporating the RiR structure into our classification approach stems from its ability to disentangle complex linguistic structures especially when both reduplication and repetition are present in the same sentence. We believe that the following examples illustrate the significance of considering the RiR structure:

The examples given below follow the standard scheme:

*[ reparandum + {interregnum} repair ]*

| Model | Hindi | | | Telugu | | | Marathi | | | avg. F1 |
|---|---|---|---|---|---|---|---|---|---|---|
| | **P** | **R** | **F1** | **P** | **R** | **F1** | **P** | **R** | **F1** | |
| *Baseline Models* | | | | | | | | | | |
| Logistic Regression | 29.76 | 18.20 | 22.59 | 28.82 | 14.95 | 19.69 | 24.52 | 18.39 | 21.02 | 21.10 |
| BiLSTM-CRF | 52.51 | 58.10 | 55.16 | 53.67 | 47.54 | 50.42 | 61.44 | 45.28 | 52.14 | 52.57 |
| *Comparison of multilingual transformer model performance with and without RiR structure* | | | | | | | | | | |
| bert-base-multilingual | 81.30 | 75.57 | 78.33 | 76.08 | 75.74 | 75.91 | 77.54 | 76.75 | 77.14 | 77.13 |
| bert-base-multilingual + RiR | 84.24 | 77.52 | 80.74 | 82.45 | 74.88 | 78.48 | 85.47 | 74.80 | 79.78 | **79.67** ↑ |
| XLMR-base | 85.18 | 80.30 | 82.67 | 84.16 | 75.19 | 79.42 | 89.67 | 74.27 | 81.25 | 81.11 |
| XLMR-base + RiR | 95.41 | 74.06 | 83.39 | 86.12 | 75.60 | 80.52 | 93.04 | 73.12 | 81.89 | **81.93** ↑ |
| XLMR-large | 84.44 | 86.32 | 85.37 | 94.44 | 75.19 | 83.72 | 88.92 | 80.51 | 84.51 | 84.53 |
| XLMR-large + RiR | 89.33 | 82.21 | 85.62 | 89.60 | 78.97 | 83.95 | 85.49 | 84.16 | 84.82 | **84.80** ↑ |

Table 4: Complete results across languages for baseline models and RiR models. Precision (P), Recall (R) and F1-score (F1) for reduplication, repetition, and other predictions at word level, including the Overall macro F1-score averaged over 5 runs are mentioned. The best results are in bold.

### 5.2.1 Examples

- **Example 2:**

  वह [नीला + { नहीं } नीला नीला] फूल है।
  vah [ neela + *nahi* + neela neela] phool hai.
  It [blue + *no* + blue blue] flower is.

  Translation: "It is a blue, no blue-blue flower."

  The interregnum "नहीं" (no) indicates repetition between first नीला (neela) and second first नीला (neela) with a negation marker as interregnum, making it distinct from reduplication between second नीला (neela) and third नीला (neela).

- **Example 3:**

  वह [नीला + { } नीला] फूल है।
  vah [ neela +  + neela] phool hai.
  It [blue + { } + blue] flower is.

  Translation: "It is a blue, no blue-blue flower."

  The empty, interregnum suggests reduplication, where the Reparandum is repeated without any additional elements.

### 5.3 Modeling Approach

To classify reduplication and repetition, we propose a model that takes into account the RiR structure. Our feature extraction process will involve capturing information from the Reparandum, Interregnum, and Repair segments. To do so, we provide the model as separate features, the words surrounding the reduplicated words. These are highlighted in green, in the above examples. This helps especially in intricate cases, where both phenomena overlap as in Example 2 in Section 5.2.1. By leveraging the RiR structure and considering the surrounding words, we believe our approach will contribute to better utilization of this structural information for the classification of reduplication and repetition.

## 6 Experimental Setup

This section details the methodology adopted to distinguish between reduplication, repetition, and other phenomena in speech transcripts.

### 6.1 Data Processing

Speech transcripts were preprocessed by removing punctuation to ensure consistency in the dataset. This step also made the task more challenging and realistic.

### 6.2 Baseline Models

We evaluated performance using two established baseline models: Logistic Regression and BiLSTM-CRF, chosen for their common use in sequence labeling and classification tasks within NLP. Logistic Regression provides a basic measure of linear separability, while BiLSTM-CRF, better suited for sequential dependencies, offers improved performance in sequence-to-sequence predictions (Huang et al., 2015).

### 6.3 Transformer-Based Models

Given the success of transformer-based models in NLP, we employed them for our task. We selected multilingual transformer architectures suitable for Hindi.

We used `bert-base-multilingual` and its variant with the Reparandum-Interregnum-Repair (RiR) structure, which are recognized for their robust language modeling capabilities. Both models were fine-tuned for our task.

Further, we explored the XLMR models: `XLMR-base`, `XLMR-base` with RiR, `XLMR-large`, and `XLMR-large` with RiR, to evaluate their performance and the possible advantages of the RiR structure.

### 6.4 Training and Finetuning Setup

Models were trained using a batch size of 8 for a maximum of 5 epochs. The `AdamW` optimizer was used with a learning rate of `1e-5`. Models were fine-tuned on a dataset specific to reduplication and repetition.

## 7 Results and Analysis

In this section, we thoroughly analyse all experiments on reduplication-repetition classification. Results for all the models across the three languages, fine-tuned on the IndicRedRep dataset, are shown in Table 4. For a more detailed examination of the language-wise results, readers can refer to Appendix A. This section also discusses some qualitative examples across languages as given in Table 5, providing interesting insights of confusion cases, and our analysis of how RiR modelling helps in resolving these cases.

To evaluate the performance of all models used in our experiments, we use precision, recall, and F1 score; metrics that are commonly used across classification tasks in natural language processing (Manning and Schutze, 1999; Jurafsky, 2000). These metrics have also been used in previous related works focusing on disfluency detection and similar tasks (Jamshid Lou and Johnson, 2017; Passali et al., 2022).

### 7.1 Baseline Models

Results in Table 4 show average F1 scores of 21.10 for Logistic Regression and 52.57 for BiLSTM-CRF, highlighting the latter's superiority in handling complex linguistic tasks. Analysis across Hindi, Telugu, and Marathi indicated superior performance in Hindi, attributed to better data resources, whereas Telugu and Marathi posed additional challenges due to their linguistic complexities.

### 7.2 Multilingual Transformer Models

We observed that fine-tuning pre-trained models on the IndicRedRep test set, specifically for detecting and identifying reduplication and repetition, yielded significantly higher accuracy compared to baseline Logistic Regression and BiLSTM-CRF models. Incorporating the Reparandum-Interregnum-Repair (RiR) structure into these multilingual transformer models further enhanced their performance, as detailed in Table 4. Specifically, models employing the RiR structure achieved superior results over standard models trained on the same dataset.

Our experiments highlighted a notable increase in performance metrics with the RiR structure. For example, the bert-base-multilingual model saw its average F1 score improve from 77.13% to 79.67% with RiR, and similar enhancements were noted with the XLMR models: the F1 score for the XLMR-base model rose from 81.11

This improvement was not uniform across all languages, reflecting the varied complexities and characteristics of Hindi, Telugu, and Marathi, which underscores the nuanced challenges of language-specific processing in NLP. Further qualitative analysis on the impact of RiR structure integration is elaborated in Section 7.3.

### 7.3 Qualtitaitve Analysis

Table 5 presents a detailed examination of specific inference cases from our model, which was applied to unseen test sentences across Hindi, Telugu, and Marathi. It highlights some consistent misclassifications that are crucial for understanding its limitations and illustrating how RiR modeling contributes to improvement.

For example, in the first row featuring a Hindi reduplication type error, घर (ghar, 'house') is misclassified as repetition. This error may be due to the model's oversensitivity to the presence of another repetition instance in the same sentence. A similar pattern of errors is observed in the Telugu and Marathi examples within Table 5. The RiR modeling approach enhances focus on the local context of the word, resulting in correct classification when the XLMR-base + RiR model is employed. The misclassifications in other examples can be explained along similar lines, underscoring the effectiveness of RiR modeling in improving the accuracy of linguistic phenomenon classification.

| Lang | Type of Error | Sentence | Prediction | Comments |
|---|---|---|---|---|
| Hi | Reduplication | को घर घर ये सेवा पहुँचे तो इसके माध्यम से मैं ये बताना चाहता हूँ की हमारे जो झारखण्ड झारखण्ड | को घर घर ये सेवा पहुँचे तो इसके माध्यम से मैं ये बताना चाहता हूँ की हमारे जो झारखण्ड झारखण्ड | घर (ghar, 'house') is an example of reduplication class, but is confused with repetition. |
| Hi | Repetition | यह हमारे समाज के लिए नहीं बल्कि प्राचीन समय समय से ही हमारा समाज जूझ रहा है अगर हमारे समाज में कहीं भी कोई घरेलु हिंसा होती है तो इसका शिकार महिलाओं को ही | यह हमारे समाज के लिए नहीं बल्कि प्राचीन समय समय से ही हमारा समाज जूझ रहा है अगर हमारे समाज में कहीं भी कोई घरेलु हिंसा होती है तो इसका शिकार महिलाओं को ही | समय (samay, 'time') is an example of repetition, but incorrectly predicted as reduplication. |
| Te | Reduplication | మరల మరల సహాయం చేసినందుకు ధన్యవాదాలు ధన్యవాదాలు | మరల మరల సహాయం చేసినందుకు ధన్యవాదాలు ధన్యవాదాలు | మరల (marala, 'again') is incorrectly predicted as repetition, while the correct label is reduplication |
| Te | Repetition | ఉదయం ఉదయం గుడ్ మార్నింగ్ చెప్పాలి ఎందుకంటే నా నమ్మకం పిల్లలు పిల్లలు తొలి పాఠశాల ఇల్లే ఇల్లే ఉంటుంది మరియు ఉపాధ్యాయురాలు తల్లే అవుతుంది. | ఉదయం ఉదయం గుడ్ మార్నింగ్ చెప్పాలి ఎందుకంటే నా నమ్మకం పిల్లలు పిల్లలు తొలి పాఠశాల ఇల్లే ఇల్లే ఉంటుంది మరియు ఉపాధ్యాయురాలు తల్లే అవుతుంది. | ఇల్లే (ille, 'house') is predicted as reduplication by the model, but it is an example of repetition. |
| Mr | Reduplication | दूध विकत असताना रुग्णालयाच्या विभागासाठी वेगवेगळी वेगवेगळी तारीख ठरवली आहे. | दूध विकत असताना रुग्णालयाच्या विभागासाठी वेगवेगळी वेगवेगळी तारीख ठरवली आहे. | वेगवेगळी (vegvegali, 'different') is an example of reduplication in Marathi which is incorrectly predicted as repetition. |
| Mr | Repetition | सतरा आदि आदि जिल्ह्यांमधून दोनशे दोनशे कार्यकर्ते सहभागी होतील, धन्यवाद. | सतरा आदि आदि जिल्ह्यांमधून दोनशे दोनशे कार्यकर्ते सहभागी होतील, धन्यवाद. | आदि (ādi, 'so on') is an example of repetition in Marathi, but it is mis-classified as reduplication. |

Table 5: Inference examples from RiR models for cases where the baseline model XLMR-base failed, but XLMR-base + RiR predicted correctly. Language codes are Hi-Hindi, Te-Telugu, and Mr-Marathi. In the prediction column, the black-colored text stands for the 'O' (no label) class, while blue-colored text stands for reduplication class prediction, and red color stands for repetition class prediction. **The translations and gloss of these sentences are given in the Appendix D.**

## 8 Conclusions and Future Work

Our study introduced and validated a model that employs the Reparandum-Interregnum-Repair (RiR) structure to enhance the classification of linguistic phenomena such as reduplication and repetition in multilingual contexts. Our experiments, as detailed in the table, demonstrated that incorporating the RiR structure consistently improves the performance across multiple languages (Hindi, Telugu, and Marathi), as evidenced by higher F1 scores when compared to baseline models, including both traditional ML-based and advanced transformer models.

The RiR structure's utility in distinguishing complex linguistic patterns is particularly notable. This approach provided clear benefits over traditional models like Logistic Regression and BiLSTM-CRF, and even showed marked improvement over advanced models like the multilingual BERT and XLMR in their standard configurations. The most significant improvements were observed with the XLMR-large + RiR model, highlighting the effectiveness of integrating structural linguistic insights into sophisticated neural architectures for NLP tasks.

Future research should expand our approach to include more languages, especially those under-represented in NLP, and explore additional linguistic structures beyond the RiR to enhance understanding of language processing. There is also potential to integrate our RiR like models to localize context in other NLP tasks like syntactic parsing or semantic analysis, which could lead to improvements in complex language understanding systems. Moreover, applying our model's insights to real-world applications such as speech recognition could significantly enhance their effectiveness and accuracy.

## 9 Limitations

Given the complexities of disambiguating reduplication and repetition in different languages, our study, while rigorous, presents limitations that are acknowledged below:

- **Generalization across Languages**:

Our experiments were limited to three languages: Hindi, Telugu, and Marathi. The generalizability of our results to other languages, particularly non-Indo-European languages, is uncertain. Future studies should explore the application of the RiR structure in a broader linguistic context to verify its effectiveness across a wider array of language families.

- **RiR Structure**: Our approach heavily relies on the RiR structure to classify disfluencies. While effective, this method assumes that the disfluency markers and structures are consistent and distinguishable, which may not hold in more colloquial or less structured speech datasets.

- **Other Subword Representations**: Our study focused exclusively on transformer-based models (BERT and XLMR) with the addition of the RiR structure. We did not include other potent subword representations like ELMo (Peters et al., 1802) and contextual string embeddings (Akbik et al., 2018), which might offer different advantages in handling complex language phenomena. The lack of availability of these models in multiple languages restricted their inclusion in our study.

To address these limitations, future research should aim to include a more diverse set of languages and linguistic structures. Moreover, experimenting with additional subword representations and extending the RiR framework to accommodate more varied disfluency types could enhance model robustness. An exploration of the impacts of different preprocessing techniques on the model's ability to recognize and classify speech patterns accurately would also be beneficial.

## A  Detailed Results

In this section we expand Table 4 and give label-wise results for each language. Tables 6, 7, 8 contains results for Hindi, Marathi and Telugu respectively.

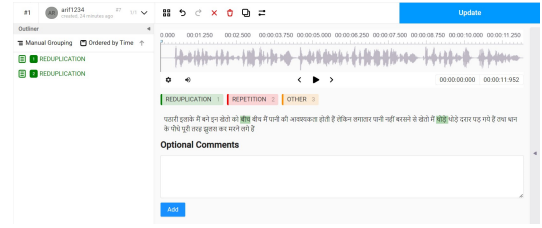

Figure 2: Interface for adding reduplication and repetition labels

## B  Annotation Guidelines

Thank you for participating in our study, on identifying reduplication and repetition in speech. During this task, you will be presented with an interface (see Fig. 2), which shows you an audio file as well as the corresponding transcript for that audio.

**Instructions**  You need to identify whether a word being repeated in the text transcript is reduplication or repetition. These are defined as below.

**Reduplication**  When we say, reduplication in this study, we mean complete reduplication. Complete reduplication, also known as full reduplication, is a linguistic process in which the entire base word is repeated to create a new word or form. In Hindi, complete reduplication is commonly used to express intensity, repetition, or to emphasize a particular action or state.

Examples of complete reduplication in Hindi sentences:

1. वे रो रहे थे, चिल्ला चिल्ला कर।
   Translation: They were crying loudly, screaming and screaming.

2. उसने धीरे धीरे सबको चुप करा दिया।
   Translation: He gradually silenced everyone.

3. वह बिलकुल बिलकुल सही था
   vah bilkul bilkul sahi tha
   Translation: He was absolutely correct.

In these examples, the complete repetition of the base word adds emphasis and intensity to the action or state described, enhancing the overall meaning of the sentences.

| Model | Reduplication | | | Repetition | | | Other | | | macro F1 |
|---|---|---|---|---|---|---|---|---|---|---|
| | P | R | F1 | P | R | F1 | P | R | F1 | |
| *Baseline Models* | | | | | | | | | | |
| Logistic Regression | 14.21 | 51.36 | 22.26 | 15.32 | 45.51 | 22.92 | 13.00 | 40.00 | 20.00 | 22.59 |
| BiLSTM-CRF | 65.41 | 51.93 | 57.90 | 62.14 | 45.32 | 52.41 | 60.00 | 42.00 | 50.00 | 55.16 |
| *Comparison of multilingual transformer model performance with and without RiR structure* | | | | | | | | | | |
| bert-base-multilingual | 81.64 | 86.44 | 83.97 | 81.84 | 83.89 | 82.85 | 62.09 | 76.99 | 68.18 | 78.33 |
| bert-base-multilingual + RiR | 83.71 | 82.86 | 83.27 | 86.18 | 85.74 | 85.96 | 69.62 | 76.99 | 73.00 | 80.74 ↑ |
| XLMR-base | 80.28 | 90.14 | 85.00 | 83.61 | 84.58 | 84.09 | 77.74 | 80.44 | 79.05 | 82.67 |
| XLMR-base + RiR | 78.86 | 92.96 | 85.33 | 91.27 | 83.37 | 87.12 | 82.77 | 73.91 | 77.74 | 83.39 ↑ |
| XLMR-large | 84.45 | 95.42 | **89.60** | 86.36 | 88.92 | 87.59 | 82.26 | 76.09 | **78.92** | 85.37 |
| XLMR-large + RiR | **88.54** | 89.79 | 89.16 | 88.48 | 92.53 | **90.46** | **87.52** | 69.57 | 77.24 | **85.62** ↑ |

Table 6: Detailed results for Hindi Language. Precision (P), Recall (R) and F1-score (F1) for reduplication, repetition, and other predictions at word level, including the Overall macro F1-score averaged over 5 runs. Best results are in bold.

| Model | Reduplication | | | Repetition | | | Other | | | macro F1 |
|---|---|---|---|---|---|---|---|---|---|---|
| | P | R | F1 | P | R | F1 | P | R | F1 | |
| *Baseline Models* | | | | | | | | | | |
| Logistic Regression | 12.23 | 50.41 | 20.37 | 13.72 | 44.67 | 20.89 | 11.45 | 39.56 | 17.81 | 19.69 |
| BiLSTM-CRF | 60.58 | 50.74 | 54.87 | 57.42 | 43.91 | 49.58 | 55.17 | 40.73 | 46.82 | 50.42 |
| *Comparison of multilingual transformer model performance with and without RiR structure* | | | | | | | | | | |
| bert-base-multilingual | 77.89 | 84.67 | 80.43 | 77.35 | 82.14 | 79.92 | 60.76 | 75.23 | 67.39 | 75.91 |
| bert-base-multilingual + RiR | 80.43 | 80.55 | 80.22 | 84.68 | 84.32 | 84.47 | 67.85 | 75.04 | 70.76 | 78.48 |
| XLMR-base | 75.96 | 88.45 | 80.63 | 80.34 | 83.27 | 81.79 | 73.58 | 78.39 | 75.84 | 79.42 |
| XLMR-base + RiR | 73.81 | 91.07 | 80.97 | 89.24 | 82.46 | 85.67 | 78.14 | 72.68 | 74.93 | 80.52 |
| XLMR-large | 82.75 | 93.39 | 87.92 | 84.15 | 87.04 | 85.67 | 80.33 | 74.87 | 77.56 | 83.72 |
| XLMR-large + RiR | 86.32 | 88.67 | 87.44 | 86.57 | 91.48 | 88.99 | 84.29 | 68.74 | 75.41 | 83.95 |

Table 7: Detailed results for Telugu Language. Precision (P), Recall (R) and F1-score (F1) for reduplication, repetition, and other predictions at word level, including the Overall macro F1-score averaged over 5 runs. Best results are in bold.

**Repetition** Repetition is a speech disfluency. Disfluencies are interruptions or disturbances that occur during speech, causing a break in the normal flow of language. Repetition, specifically word repetition, occurs when a speaker repeats a single word one or more times in their speech. This type of disfluency can happen due to hesitation, uncertainty, nervousness, lack of confidence, speech disorders, cognitive processing issues or as a natural part of the speech process.

Examples of word repetition disfluencies in Hindi:

1. मैं मैं घर जा रहा हूँ।
   (mai ghar ja rha hu)
   Translation: I-I am going home.

2. मैं घर घर जा रहा हूँ।
   Translation: I am going home home.

**Neither exists**

1. दिल की बातों उसे दे रही मात मात से कोई बनेगी नहीं बात पलायन छोड़े करें दिल की बात रेकॉर्ड बनाया था पलायन ने। (sentence boundary)

2. मैं एन सी सी का छात्र हूं। (organization short form)

3. मेरा फोन नंबर है नौ दो एक एक। (numbers)

4. के लिए आज परीक्षा आयोजित की गई जिसमें अभ्यर्थियों का प्रमाणपत्र वेरिफिकेशन लिखित परीक्षा एवं साक्षात्कार का आयोजन किया गया विद्यालय परिसर में ही किया गया इस आयोजन में लगभग साठ अभ्यर्थियों ने योगदान किया मैं राजीव कुमार ठाकुर ग्राम राइसेर पोस्ट वाजिपुर ज़िला मुंगेर मुंगेर मोबाइल वाणी से धन्यवाद।

**Instructions for transcript correction in annotation** We need to correct the speech transcripts before labelling them for reduplication and repetition as the speech transcripts have a lot of errors. To do, so we use the interface as shown in Fig. 3.

| Model | Reduplication | | | Repetition | | | Other | | | macro F1 |
|---|---|---|---|---|---|---|---|---|---|---|
| | **P** | **R** | **F1** | **P** | **R** | **F1** | **P** | **R** | **F1** | |
| *Baseline Models* | | | | | | | | | | |
| Logistic Regression | 13.67 | 52.00 | 21.58 | 14.76 | 46.82 | 22.45 | 12.59 | 41.33 | 19.04 | 21.02 |
| BiLSTM-CRF | 63.00 | 51.75 | 56.82 | 59.18 | 44.90 | 51.00 | 57.65 | 41.58 | 48.61 | 52.14 |
| *Comparison of multilingual transformer model performance with and without RiR structure* | | | | | | | | | | |
| bert-base-multilingual | 79.40 | 85.23 | 82.21 | 79.12 | 82.78 | 80.90 | 61.85 | 76.46 | 68.32 | 77.14 |
| bert-base-multilingual + RiR | 82.05 | 81.90 | 82.00 | 85.53 | 84.67 | 85.10 | 68.73 | 75.98 | 72.25 | 79.78 |
| XLMR-base | 77.89 | 89.33 | 83.11 | 82.25 | 84.42 | 83.33 | 75.58 | 79.04 | 77.31 | 81.25 |
| XLMR-base + RiR | 75.76 | 92.22 | 82.99 | 90.17 | 82.59 | 86.28 | 79.90 | 73.12 | 76.41 | 81.89 |
| XLMR-large | 83.67 | 94.21 | 88.69 | 85.22 | 88.06 | 86.63 | 81.40 | 75.25 | 78.22 | 84.51 |
| XLMR-large + RiR | 87.21 | 89.58 | 88.39 | 87.05 | 91.97 | 89.49 | 85.33 | 69.38 | 76.58 | 84.82 |

Table 8: Detailed results for Marathi Language. Precision (P), Recall (R) and F1-score (F1) for reduplication, repetition, and other predictions at word level, including the Overall macro F1-score averaged over 5 runs. Best results are in bold.

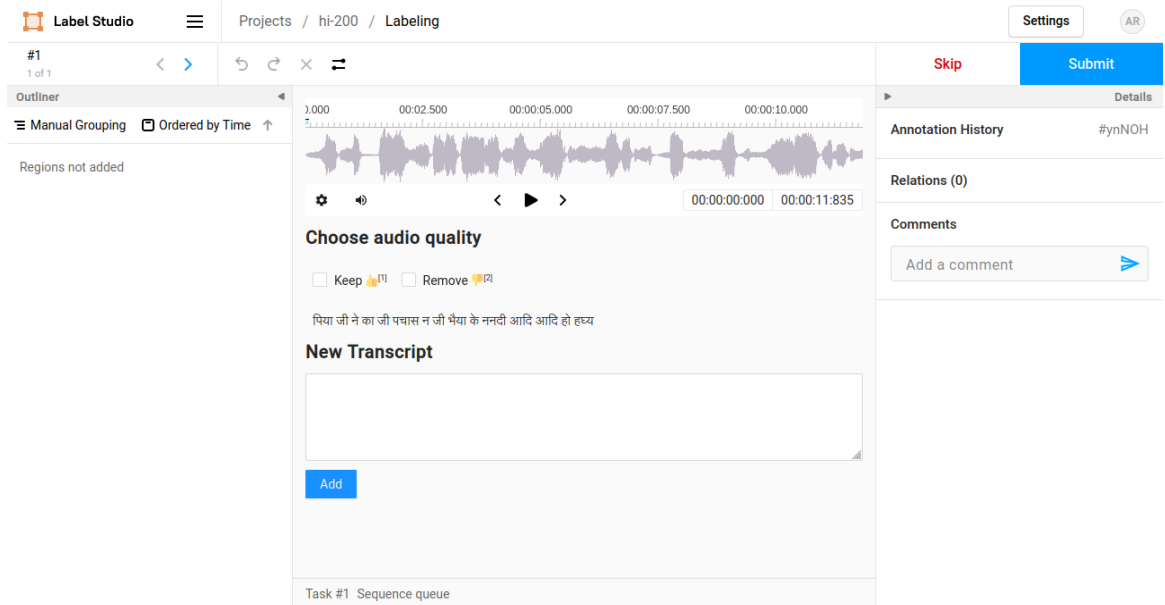

Figure 3: Interface for transcript correction of audio files

Please follow the following steps while annotating the data:

- First, copy and paste the text in the box below the title **New Transcript**

- Next, play the audio and listen to it carefully, while also reading the hindi text.

- 
  - If the hindi text is correct, then submit the simple copy paste of the hindi text as it is, checkbox the Keep button and click on the Submit Button.
  - Else, Correct the words in the box, based on the audio. Make sure to not add any punctuations like (|,-.) and also donot use any numerals in 0-9.
    * If after correcting the hindi text, you find that reduplication / repetition word is removed, then click on the Remove check box and then on the blue Submit button
    * Else, if there is still reduplication or repetition in the corrected text, click on Keep checkbox and then click on the Blue Submit button.

## C  Gemma LLM Prompting Details

This section provides the details of the prompts used to generate sentences with reduplication and repetition in Marathi and Telugu using the Gemma Instruction Tuned models. The prompts are designed to elicit specific linguistic phenomena from the model, ensuring the generated sentences closely mimic the structures observed in the Hindi subset of the IndicRedRep dataset.

The exact prompts used are listed below in a formatted box to highlight their syntactic structure and key phrases, which can be directly replicated for similar tasks.

```
### Instruction:

Reduplication is defined as a word or
    ↪ part of a word is repeated to
    ↪ convey nuances such as emphasis,
    ↪  intensity, plurality, or
    ↪ grammatical aspects.
Generate examples of natural sentences
    ↪ , that use reduplication. The
    ↪ sentence should be meaningful.

### **Examples:**
- [Telugu example 1 here]
- [Telugu example 2 here]

### **Input:**
- [Hindi sentence 1 from GV corpuse
    ↪ here]
- [Hindi sentence 2 from GV corpuse
    ↪ here]

Generate five new distinct
    ↪ reduplication sentences in
    ↪ Telugu.
### **Output:**
1.
```

## D  Qualitative analysis

The gloss for each example listed in Table 5 is given as a list in the same order in Table 9.

| |
|---|
| **Hindi:** को घर घर ये सेवा पहुँचे तो इसके माध्यम से मैं ये बताना चाहता हूँ की हमारे जो झारखण्ड झारखण्ड <br> *Transliteration:* Ko ghar ghar ye sevā pahuncē to iske mādhyam se main ye batānā chāhtā hūn ki hāmare jo Jharkhand Jharkhand <br> *Translation:* When this service reaches each home, I want to convey through this that our Jharkhand, Jharkhand... |
| **Hindi:** यह हमारे समाज के लिए नहीं बल्कि प्राचीन समय समय से ही हमारा समाज जूझ रहा है अगर हमारे समाज में कहीं भी कोई घरेलु हिंसा होती है तो इसका शिकार महिलाओं को ही <br> *Transliteration:* Yah hamāre samāj ke liye nahīn balki prāchīn samay samay se hī hamārā samāj jūjh rahā hai agar hamāre samāj mein kahīn bhī koī gharelū hiṃsā hotī hai to iskā shikār mahilāon ko hī <br> *Translation:* This is not for our society but from ancient times our society has been struggling, if there is any domestic violence anywhere in our society, then it is the women who are the victims. |
| **Telugu:** మరల మరల సహాయం చేసినందుకు ధన్యవాదాలు ధన్యవాదాలు <br> *Transliteration:* Marala marala sahāyaṁ chēsinanduku dhanyavādālu dhanyavādālu <br> *Translation:* Thank you again and again for the help. |
| **Telugu:** ఉదయం ఉదయం గుడ్ మార్నింగ్ చెప్పాలి ఎందుకంటే నా నమ్మకం పిల్లలు పిల్లలు తొలి పాఠశాల ఇల్లే ఇల్లే ఉంటుంది మరియు ఉపాధ్యాయురాలు తల్లే అవుతుంది <br> *Transliteration:* Udayaṁ udayaṁ good morning cheppāli endukaṇṭē nā nammakaṁ pillalu pillalu toli pāṭhaśāla ille ille uṇṭundi mariyu upādhyāyurālu tallē avutundi <br> *Translation:* Say good morning every morning because my belief is that children's first school is the home, home, and the teacher is the mother. |
| **Marathi:** दूध विकत असताना रुग्णालयाच्या विभागासाठी वेगवेगळी वेगवेगळी तारीख ठरवली आहे <br> *Transliteration:* Dūdh vikat asatānā rugṇālayāchyā vibhāgāsāṭhī vegvegaḷī vegvegaḷī tārīkh ṭharavalī āhe <br> *Translation:* Different dates have been set for the hospital's department while selling milk. |
| **Marathi:** सतरा आदि आदि जिल्ह्यांमधून दोनशे दोनशे कार्यकर्ते सहभागी होतील, धन्यवाद <br> *Transliteration:* Satarā ādi ādi jilhyānmadhūn donaśe donaśe kāryakarte sahbhāgī hotīl, dhanyavād <br> *Translation:* From seventeen etc., etc., districts, two hundred two hundred workers will participate, thank you. |

Table 9: Complete sentences with translations and glosses from Table 5